# Séparation des Solutions aux Modèles Géométriques Direct et Inverse pour les Manipulateurs Pleinement Parallèles

Chablat Damien, Wenger Philippe

Institut de Recherche en Communications et Cybernétique de Nantes

UMR CNRS 6597, 1 rue de la Noë B.P. 6597

44321 Nantes Cedex 3

Tél. 02.40.37.69.48, Fax. 02.40.37.69.30

Damien.Chablat@irccyn.ec-nantes.fr

**RESUME**

Cet article fournit un formalisme permettant de gérer les solutions des modèles géométriques direct et inverse des manipulateurs pleinement parallèles. Nous introduisons la notion de « modes de fonctionnement » pour séparer les solutions du modèle géométrique inverse. Puis, nous définissons, pour chaque mode de fonctionnement, les aspects de ces manipulateurs. Pour séparer les solutions du modèle géométrique direct, nous introduisons la notion de surfaces caractéristiques. Puis, nous définissons les domaines d'unicité, comme étant les plus grands domaines de l'espace de travail dans lesquels il y a unicité de solutions. Les applications principales de ce travail sont la conception, la planification de trajectoires et la commande.

# 1. INTRODUCTION

L'utilisation de manipulateurs dans l'industrie n'a cessé d'augmenter ces dernières années. Ils ont permis d'augmenter la production et la précision des mouvements effectués. La plus grande partie de ces manipulateurs est constituée des manipulateurs sériels dont la morphologie peut facilement être comparée à celle de l'être humain. Plus récemment, les manipulateurs parallèles sont apparus dans les domaines où l'on a besoin d'effectuer des mouvements rapides ou des placements très précis (simulateur de vol, usinage, chirurgie, etc.) ou transporter de lourdes charges. Un problème important non résolu lié à l'utilisation de ce type de mécanisme est la gestion des différentes solutions aux modèles géométriques.

La résolution de ce problème passe par la détermination de domaines d'unicité des opérateurs géométriques.

Historiquement, **[1]** a introduit la notion d'aspect pour les manipulateurs sériels et **[2]** a étendu cette notion pour les mécanismes cuspidaux par l'introduction des régions et composantes de base.

Dans un précédent travail, nous avons redéfini la notion d'aspect pour les manipulateurs parallèles possédant une seule solution au modèle géométrique inverse **[3]**.

Le but de cet article est de formaliser la notion d'aspect pour les manipulateurs parallèles possédant plusieurs solutions au modèle géométrique inverse. Pour ce faire, nous définissons des modes de fonctionnement, qui permettent de séparer les solutions du modèle géométrique inverse. Ainsi, nous définissons de nouveaux aspects. Nous montrons qu'il est aussi possible de définir une trajectoire de changement de mode d'assemblage, à l'intérieur d'un même aspect, exempte de toute singularité. Ce qui montre que la notion d'aspect, limitée à l'étude des singularités, n'était pas suffisante pour définir les domaines d'unicité d'un tel manipulateur. Nous définissons des nouveaux domaines d'unicité grâce à la notion de surfaces caractéristiques que nous introduisons. Nous décrivons de même un vocabulaire permettant de définir les singularités, les matrices jacobiennes et les aspects qui en découlent. Le travail présenté dans cet article est illustré à l'aide d'exemples.

## 2. PRELIMINAIRES

### 2.1. MANIPULATEURS ETUDIES

#### 2.1.1. Définition 1 : Les manipulateurs pleinement parallèles

Ce sont des mécanismes comportant autant de chaînes cinématique élémentaires que l'effecteur présente de degré de liberté, chaque chaîne cinématique élémentaire comporte au plus un actionneur. De plus, aucun segment d'une chaîne cinématique élémentaire ne doit être relié à plus de deux corps (chaîne cinématique série) **[4]**. Ces manipulateurs sont donc non redondants et nous supposons les chaînes cinématiques élémentaires indépendantes.

**Définition 2 : Une chaîne cinématique élémentaire**

En parlant d'un manipulateur parallèle, c'est une chaîne cinématique reliant la base fixe à l'effecteur du manipulateur. On appelle aussi celle-ci, jambe d'un manipulateur **[5]**.

### 2.2. LES ASPECTS

#### 2.2.1. Espace articulaire et espace opérationnel

On peut définir une relation entre les entrées (**q**), c'est-à-dire les variables articulaires et les sorties (**X**), c'est-à-dire les configurations de la plate-forme mobile de la manière suivante :

$$F(\mathbf{X}, \mathbf{q}) = 0 \tag{1}$$

Cette définition peut être appliquée à tout type de manipulateur, qu'il soit sériel ou parallèle. On note :

- $EA_n$ l'espace articulaire lié aux articulations motorisées **q** ($n$ désigne le nombre d'articulations motorisées) ;
- $EAP_r$ l'espace articulaire lié aux articulations passives $\mathbf{q}_p$ ($r$ désigne le nombre d'articulations passives) ;
- $EO_m$ l'espace opérationnel lié aux configurations de la plate-forme mobile ($m$ désigne le nombre de degré de liberté de la plate-forme mobile).

L'ensemble articulaire $Q$ est l'ensemble des positions articulaires motorisées permettant d'assembler le manipulateur.

- $Q \subset EA_n$
- $Q = \left\{ \begin{array}{l} \mathbf{q} \in EA_n / \mathbf{q}_p \in EAP_p, \begin{array}{ll} \forall\ i \leq n & q_{i\,min} \leq q_i \leq q_{i\,max} \\ \forall\ j \leq r & q_{p\,j\,min} \leq q_{p\,j} \leq q_{p\,j\,max} \end{array} \\ \exists \mathbf{X} \in EO_m / F(\mathbf{X}, \mathbf{q}) = 0 \end{array} \right\}$

L'espace de travail *W* est l'ensemble des positions et des orientations accessibles par un repère lié à l'organe effecteur du manipulateur défini tel que :

- $W \subset EO_m$
- $\forall \mathbf{X} \in W$ et $\forall \mathbf{q} \in Q$, $F(\mathbf{X}, \mathbf{q}) = 0$

### 2.2.2. Les singularités

Pour identifier les configurations singulières, nous utilisons la méthode proposée par Gosselin [**14**]. Par différentiation de l'équation (1), nous obtenons l'équation suivante :

$$\mathbf{A}\,\mathbf{t} + \mathbf{B}\,\dot{\mathbf{q}} = 0 \qquad (2)$$

Avec :

- $\mathbf{t} \equiv [\varpi, \dot{\mathbf{c}}]$ dans le cas des manipulateurs plans, où $\varpi$ représente la vitesse de rotation plane et $\dot{\mathbf{c}}$ la vitesse de déplacement dans le plan ;

- $\mathbf{t} \equiv [\varpi]$ dans le cas des manipulateurs sphériques, où $\varpi$ représente le vecteur rotation ;

- $\mathbf{t} \equiv [\varpi, \dot{\mathbf{c}}]$ dans le cas des manipulateurs spatiaux, où $\varpi$ représente le vecteur rotation et $\dot{\mathbf{c}}$ la vitesse de déplacement dans l'espace ;

- $\dot{\mathbf{q}} = [\dot{q}_1, \dot{q}_2, .., \dot{q}_m]^T$ représente le vecteur des vitesses articulaires liées aux articulations motorisées.

- $\mathbf{A}$ et $\mathbf{B}$ sont deux matrices jacobiennes.

On appelle singularité toute configuration telle que $\mathbf{A}$ ou $\mathbf{B}$ (ou les deux) n'est (ne sont) plus inversible(s). Nous pouvons ainsi étudier 3 types de configurations singulières :

a) Det($\mathbf{A}$) = 0

b) Det($\mathbf{B}$) = 0

c) Det($\mathbf{A}$) = 0 et Det($\mathbf{B}$) = 0

**Remarque**

Dans la plupart des études réalisées sur les manipulateurs pleinement parallèles, on écrit souvent l'équation (2) sous la forme $\mathbf{t} = \mathbf{J}\,\dot{\mathbf{q}}$ avec $\mathbf{J} = \mathbf{A}^{-1}\mathbf{B}$. Ce formalisme, emprunté aux manipulateurs sériels, ne permet pas de distinguer les différents types de configurations singulières. De plus, cette notation utilise implicitement l'inversion de la matrice jacobienne parallèle. Or celle-ci n'est pas toujours inversible.

**Définition 3 : Les singularités parallèles**

Les singularités parallèles sont dues à la perte de rang de matrice jacobienne **A** que l'on nommera par la suite matrice jacobienne parallèle, c'est-à-dire lorsque Det(**A**) = 0. Dans ce cas, il est possible de déplacer la plate-forme mobile alors que les articulations motorisées sont bloquées. Dans ces configuration, le manipulateur gagne un ou plusieurs degré(s) de liberté. Ces singularités sont propres aux manipulateurs parallèles.

**Définition 4 : Les singularités sérielles**

Les singularités sérielles sont dues à la perte de rang de matrice jacobienne **B** que l'on nommera par la suite matrice sérielle, c'est-à-dire lorsque Det(**B**) = 0. Dans ce cas, il n'est pas possible d'engendrer certaines vitesses de la plate-forme mobile. Les singularités sérielles représentent les limites de l'espace de travail **[6]**. Dans ces configurations, le manipulateur perd un ou plusieurs degré(s) de liberté.

### 2.2.3. Notion d'aspect

La notion d'aspect a été introduite pour des manipulateurs possédant les caractéristiques suivantes :

a) Manipulateur sériel **[1]** ;
b) Manipulateur parallèle avec plusieurs solutions au modèle géométrique direct et une seule solution au modèle géométrique inverse **[7] [3]** ;

**Définition 5 : Les aspects parallèles [3]**

Nous appelons aspects parallèles, les aspects définis pour les manipulateurs parallèles possédant plusieurs solutions au modèle géométrique direct et une seule solution au modèle géométrique inverse.

Un aspect parallèle (noté *WA*) est un domaine de l'espace de travail *W* où le déterminant de la matrice jacobienne parallèle **A** n'est jamais nul, sauf s'il est nul partout sur **A** :

- *WA* ⊂ *W* ;
- *WA* est connexe ;
- ∀ **X** ∈ WA , Det(**A**) ≠ 0.

Les aspects parallèles sont donc les plus grands domaines de *W* dépourvus de toute singularité parallèle (Définition 3). Les aspects parallèles séparent l'espace de travail. Cette définition est insuffisante pour définir la notion d'aspect dans le cas des manipulateurs pleinement parallèles possédant plusieurs solutions au modèle géométrique inverse. Il faut la généraliser.

**Définition 6 : Les aspects introduits pour les manipulateurs sériels [1]**

La notion d'aspect a été introduite pour les mécanismes sériels. Un aspect (noté *QA*) est un domaine de l'ensemble articulaire *Q* où le déterminant de la matrice jacobienne **J** (**J** est la matrice jacobienne des manipulateurs sériels) n'est jamais nul, sauf s'il est nul partout sur **QA** :

- $QA \subset Q$ ;
- *QA* est connexe ;
- $\forall \mathbf{q} \in QA$, $\text{Det}(\mathbf{J}) \neq 0$.

Les aspects sont donc les plus grands domaines de *Q* dépourvus de toute singularité sérielle (Définition 4). Les aspects séparent l'ensemble articulaire. Nous allons tester cette définition pour caractériser les aspects des manipulateurs parallèles dus aux singularités sérielles.

## 2.3. CHANGEMENT DE POSTURE ET DE MODE D'ASSEMBLAGE

### 2.3.1. Changement de posture

Les différentes postures d'un manipulateur parallèle sont dues à l'existence de plusieurs solutions au modèle géométrique inverse. Nous associons ainsi plusieurs *postures* pour la même position et la même orientation de l'effecteur du manipulateur (Figure 1). Un changement de posture revient à changer de solution du modèle géométrique inverse.

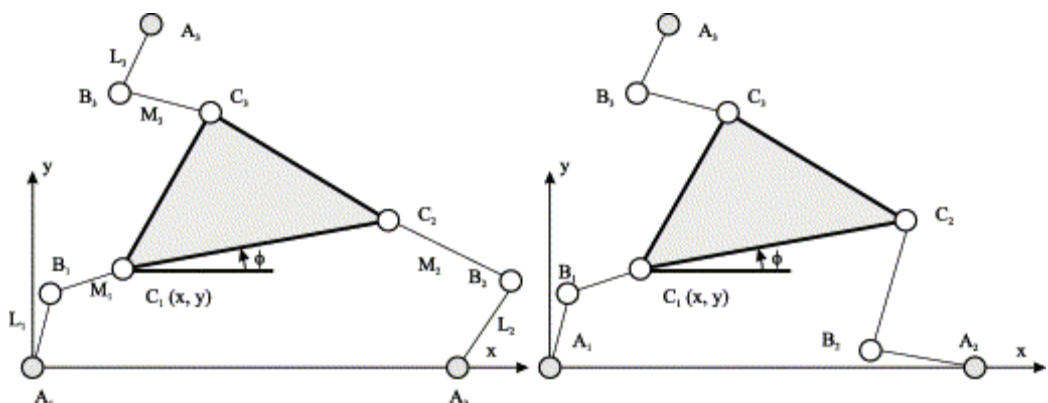

**Figure 1 : Exemple de changement de posture**

### 2.3.2. Changement de mode d'assemblage

Les différents modes d'assemblage d'un manipulateur parallèle sont dus à l'existence de plusieurs solutions au modèle géométrique direct. Il est ainsi possible d'associer plusieurs *modes* *d'assemblage* pour les mêmes coordonnées articulaires **[6]** (Figure 2). Un changement de mode d'assemblage revient à changer de solution du modèle géométrique direct. Dans le cas du manipulateur parallèle plan de type 3-R<u>P</u>R, **[8]** et **[3]** montrent qu'il est possible de changer de mode d'assemblage sans passer par une singularité parallèle.

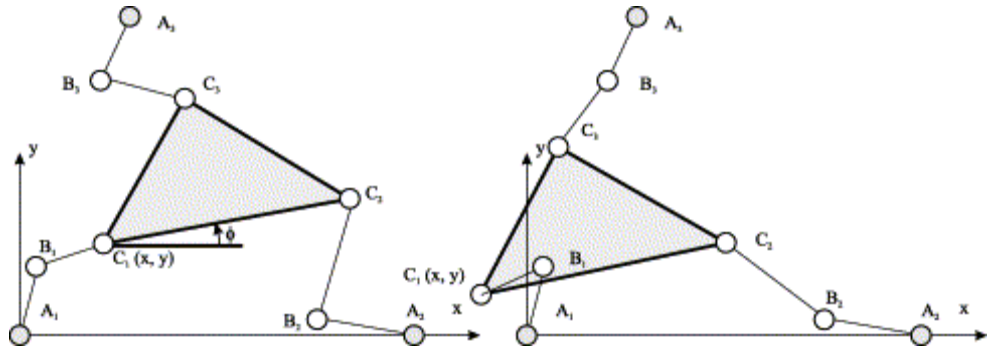

**Figure 2 : Changement de mode d'assemblage**

### 2.4. LES MODES DE FONCTIONNEMENT

Nous définissons les modes de fonctionnement pour les manipulateurs pleinement parallèles (Définition 1). Ces manipulateurs possèdent des jambes indépendantes comportant les éléments suivants : prismatique, pivot, cardan et rotule.

De cette propriété, nous pouvons en déduire la forme de la matrice jacobienne sérielle **B**, pour un manipulateur à *n* degré de liberté (Équation 3).

$\mathbf{B} = \text{Diag}[B_{11}, .., B_{jj}, .., B_{nn}]$  (3)

A chaque terme $B_{jj}$ est associé une jambe du manipulateur. Son annulation provoque l'apparition d'une singularité sérielle.

**Définition 7 : Les modes de fonctionnement**

Un mode de fonctionnement, noté $Mf_i$ est l'ensemble des configurations du mécanisme (**X**, **q**) pour lesquelles $B_{jj}$ (j = 1 à n) ne change pas de signe et ne s'annule pas. Soit :

$$Mf_i = \left\{ (\mathbf{X}, \mathbf{q}) \in W \times Q \setminus \text{signe}(B_{jj}) = \text{constantes pour } j = 1 \text{ à } n \text{ et } \det(\mathbf{B}) \neq 0 \right\}$$
(4)

L'ensemble des modes de fonctionnement $Mf = \{Mf_i\}$, $i \in I$ est donc obtenu en utilisant toutes les permutations de signe de chaque terme $B_{jj}$. S'il existe des contraintes sur les articulations (limites articulaires ou collisions internes) alors les modes de fonctionnement ne sont pas toujours tous accessibles.

Dans le théorème 1, nous montrons que les modes de fonctionnement permettent de séparer les solutions du modèle géométrique inverse dans le cas où les jambes ne sont pas cuspidales[1]. C'est-à-dire que les solutions sont bien séparées par les singularités sérielles. On peut trouver dans **[10]** la liste des chaînes sérielles non cuspidales les plus courantes. Changer de mode de fonctionnement revient à effectuer un changement de posture pour l'une des jambes.

**Théorème 1 :**

Les modes de fonctionnement séparent les solutions du modèle géométrique inverse si et seulement si les jambes du manipulateur sont non cuspidales.

**Démonstration :**

Si une jambe du manipulateur est cuspidale alors le manipulateur peut exécuter une trajectoire de changement de posture sans rencontrer de configuration singulière. Dans ce cas, aucun terme de la matrice sérielle **B** ne s'annule. Réciproquement, si aucune des jambes n'est cuspidale alors une trajectoire de changement de posture implique obligatoirement de passer par une ou plusieurs configuration(s) singulière(s) où le déterminant de la matrice jacobienne sérielle **B** s'annule.

Pour un mode de fonctionnement $Mf_i$, nous n'avons qu'une seule solution au modèle géométrique inverse du manipulateur, nous pouvons donc définir une application permettant d'associer une coordonnée opérationnelle **X** à une coordonnée articulaire **q**.

$$g_i(\mathbf{X}) = \mathbf{q} \qquad (5)$$

De même, nous posons la notation suivante permettant de définir l'ensemble des images dans $W$ d'une posture donnée de $Q$.

$$g_i^{-1}(\mathbf{q}) = \left\{ \mathbf{X} \setminus (\mathbf{X}, \mathbf{q}) \in Mf_i \right\} \qquad (6)$$

Où l'indice $i$ représente l'indice associé au mode de fonctionnement $Mf_i$.

---

[1] Une chaîne sérielle est dite cuspidale s'il existe une trajectoire de changement de posture qui ne passe pas par une singularité [**9**]

**Exemples :**

Dans le cas du *robot Delta* de **[11]**, manipulateur à trois degrés de liberté, il existe 8 modes de fonctionnement (3 jambes avec 2 postures par jambe, avec $2^3 = 8$ modes de fonctionnement). Les singularités sérielles apparaissent lorsque les articulations des chaînes cinématique élémentaires sont alignées. Et pour le *robot Hexa* de **[12]**, manipulateur à six degrés de liberté, il existe 64 modes de fonctionnement ($2^6 = 64$).

## 3. EXTENSION DE LA NOTION D'ASPECT

### 3.1. INTRODUCTION

La définition (5) permet de définir les aspects pour les manipulateurs parallèles possédant une seule solution au modèle géométrique inverse. La définition (6) permet de définir les aspects pour les manipulateurs sériels.

Pour étendre ces définitions aux manipulateurs parallèles ayant plusieurs solutions au modèle géométrique inverse et au modèle géométrique direct, nous illustrons notre propos avec un manipulateur plan à deux degrés de liberté de type RR-RRR (Figure 3).

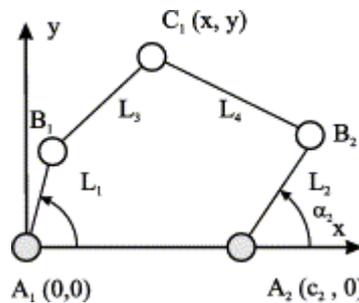

**Figure 3 : Manipulateur à boucle fermée plan de type RR-RRR**

Ce mécanisme comporte seulement une boucle fermée et l'organe effecteur est placé en $C_1$. Les pivots $A_1$ et $A_2$ sont des articulations motorisées alors que les pivots $B_1$, $B_2$ et $C_1$ sont des articulations passives. Nous avons choisi d'étudier le mécanisme possédant les dimensions $L_1 = 8$, $L_2 = 5$, $L_3 = 5$, $L_4 = 8$, $c_1 = 0$ et $c_2 = 9$.

### 3.2. ÉTUDE D'UN MANIPULATEUR PLAN

L'étude de la cinématique de ce mécanisme nous amène à la résolution du modèle géométrique direct et du modèle géométrique inverse. Nous pouvons facilement écrire le système (7) pour décrire le fonctionnement de ce mécanisme.

$$\{ (x - L_1 \cos(\theta_1))^2 + (y - L_1^2 \sin(\theta_1))^2 = L_3^2,\ (x - L_2 \cos(\theta_2) - c_2)^2 + (y - L_2 \sin(\theta_2))^2 = L_4^2 \quad (7)$$

### 3.2.1. Modèle géométrique inverse

Nous pouvons découpler les deux équations du système (7) pour obtenir les solutions en $\alpha_1$ et $\alpha_2$.

$$x^2 - 2xL_1\cos(\theta_1) + y^2 - 2yL_1\sin(\theta_1) + L_1^2 - L_3^2 = 0 \tag{8a}$$

$$x^2 - 2x(L_2\cos(\theta_2) + c_2) + L_2c_2\cos(\theta_2) + y^2 + c_2^2 - 2yL_2\sin(\theta_2) + L_2^2 - L_4^2 = 0 \tag{8b}$$

Nous obtenons ainsi deux équations du second degré. Le nombre maximum de solutions est donc de quatre. Ces deux équations étant indépendantes, nous aurons au maximum 2 racines doubles.

### 3.2.2. Modèle géométrique direct

Le modèle géométrique direct possède 2 solutions placées symétriquement par rapport à l'axe $B_1B_2$. À partir du système d'équations (7), nous écrivons le polynôme du second degré suivant :

$$4(a^2 + b^2)Y^2 - 4bcY - 4a^2L_3^2 + c^2 = 0 \tag{9}$$

Avec :

$$X = x - L_1\cos(\theta_1), \; Y = y - L_1\sin(\theta_1)$$

$$a = L_1\cos(\theta_1) - L_2\cos(\theta_2) - c_2, \; b = L_1\sin(\theta_1) - L_2\sin(\theta_2), \; c = L^2{}_4 - L^2{}_3 - b^2 - a^2$$

Les racines de l'équation 9 sont les solutions du modèle géométrique direct.

### 3.2.3. Singularités

À partir de l'équation (2), nous pouvons définir les deux matrices jacobiennes du manipulateur.

$$\mathbf{A} = \begin{bmatrix} x - L_1\cos(\theta_1), \; y - L_1\sin(\theta_1) & x - L_2\cos(\theta_2) - c_2 \\ & y - L_2\sin(\theta_2) \end{bmatrix} \tag{10a}$$

$$\mathbf{B} = \begin{bmatrix} L_1\sin(\theta_1)(x - L_1\cos(\theta_1)) \\ -L_1\cos(\theta_1)(y - L_1\sin(\theta_1)) & 0 \\ & L_2\sin(\theta_2)(x - L_2\cos(\theta_2) - c_2) \\ 0 & -L_2\cos(\theta_2)(y - L_2\sin(\theta_2)) \end{bmatrix} \tag{10b}$$

**Les singularités parallèles**

Les singularités parallèles (Définition 3) représentent les limites de l'ensemble articulaire.

$$\det(\mathbf{A}) = (x - L_1\cos(\theta_1))(y - L_2\sin(\theta_2)) - (x - L_2\cos(\theta_2))(y - L_1\sin(\theta_1))$$

Soit les vecteurs suivants :

$$\overrightarrow{B_iC_1} = (x - L_i\cos(\theta_i) - c_i, \; y - L_i\sin(\theta_i)) \qquad \text{pour } i = 1,2$$

$\det(\mathbf{A})=0$ si $\overrightarrow{B_1C_1} \wedge \overrightarrow{B_2C_1} = \vec{0}$, c'est-à-dire lorsque les points $B_1C_1B_2$ sont alignés (Figure 4).

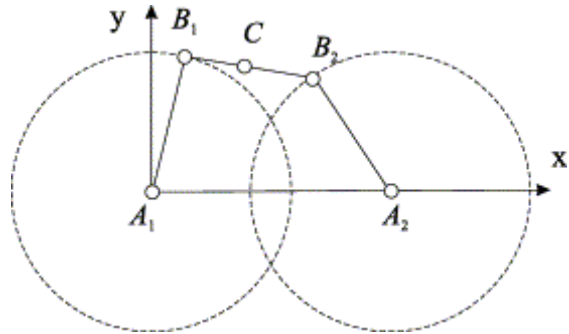

**Figure 4 : Exemple de singularité parallèle**

**Les singularités sérielles**

Pour connaître les singularités sérielles (Définition 4) nous utilisons une approche géométrique pour expliquer leurs positions. Soit les vecteurs suivants :

$$\overrightarrow{A_iB_i} = \left(L_i\cos(\theta_i), L_i\sin(\theta_i)\right) \quad \text{et} \quad \overrightarrow{B_iC_i} = \left(x - L_i\cos(\theta_i) - c_i, y - L_i\sin(\theta_i)\right) \quad \text{pour } i = 1, 2$$
avec $c_1 = 0$

$$\det(\mathbf{B}) = 0 \text{ si } \begin{cases} L_1\sin(\theta_1)(x - L_1\cos(\theta_1)) - L_1\cos(\theta_1)(y - L_1\sin(\theta_1)) = 0 \\ \text{ou} \\ L_2\sin(\theta_2)(x - L_2\cos(\theta_2) - c_2) - L_2\cos(\theta_2)(y - L_2\sin(\theta_1)) = 0 \end{cases} \quad (11)$$

Soit si $\overrightarrow{A_iB_i} \wedge \overrightarrow{B_iC_i} = \overrightarrow{0}$, c'est-à-dire lorsque les points $A_iB_iC_1$ sont alignés (Figure 5).

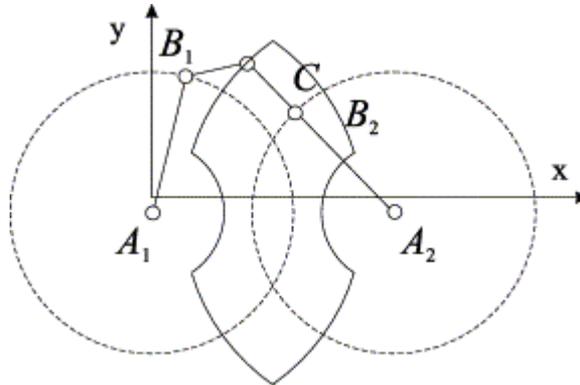

**Figure 5 : Exemple de singularité sérielle**

**Calcul des singularités sérielles**

A partir du résultat précédent, nous pouvons déterminer les conditions géométriques nécessaires à la détermination des différentes configurations singulières. Nous pouvons dénombrer 4 types de singularités :

a) Alignement de $A_1B_1C_1$ avec $\|\overrightarrow{A_1C_1}\| = L_1 + L_3$

b) Alignement de $A_1B_1C_1$ avec $\|\overrightarrow{A_1C_1}\| = L_1 - L_3$

c) Alignement de $A_2B_2C_1$ avec $\|\overrightarrow{A_2C_1}\| = L_2 + L_4$

d) Alignement de $A_2B_2C_1$ avec $\|\overrightarrow{A_2C_1}\| = L_2 - L_4$

Dans ces configurations, nous pouvons exprimer $\theta_2$ en fonction de $\theta_1$ ou $\theta_1$ en fonction de $\theta_2$.

## 3.3. LES ASPECTS : ETUDE D'UN EXEMPLE

### 3.3.1. Utilisation de la notion d'aspect parallèle (Définition 5)

Le calcul des aspects parallèles repose sur l'utilisation de la définition 5. Dans le cas du manipulateur RR-RRR, il est possible d'associer deux modes d'assemblage, les deux solutions du modèle géométrique direct. Nous obtenons deux ensembles, l'un où le déterminant de matrice jacobienne parallèle est positif (Figure 6) et l'autre où il est négatif (Figure 7). On remarque que les ensembles sont identiques tant dans l'ensemble articulaire que dans l'espace de travail. Les singularités parallèles sont les limites de l'ensemble articulaire. Pour ces configurations, il n'existe qu'une solution double au modèle géométrique direct. La position de l'image de ces configurations dans l'espace de travail est identique. Cependant, pour une configuration donnée du point $C_1$, il peut exister des configurations du manipulateur qui soient en configurations singulières et d'autres en configurations régulières. En effet, pour chaque configuration du point $C_1$, il existe 4 solutions au modèle géométrique inverse.

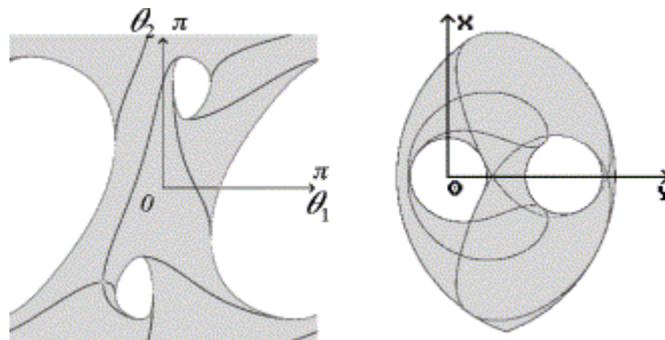

**Figure 6 : Étude du manipulateur RR-RRR pour un déterminant de la matrice jacobienne parallèle positif**

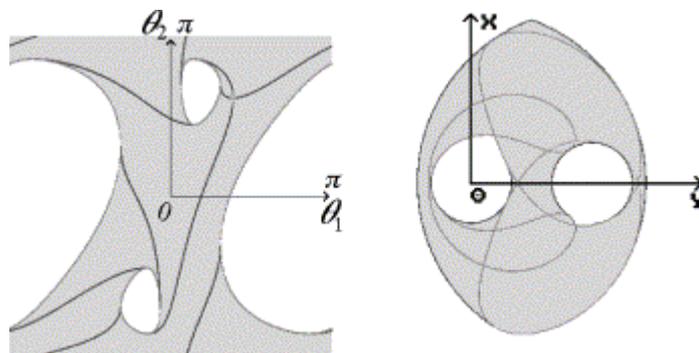

**Figure 7 : Étude du manipulateur RR-RRR pour un déterminant de la matrice jacobienne parallèle négatif**

En conclusion, l'utilisation du signe du déterminant de la matrice jacobienne parallèle **A** ne permet pas de décomposer l'espace de travail en aspects. Nous devons donc rechercher une autre formulation de la notion d'aspect pour résoudre ce problème.

### 3.3.2. Utilisation de la notion d'Aspect introduite pour les manipulateurs sériels (Définition 6)

Nous allons tester la définition de la notion au sens de Borrel (Définition 6). La valeur du déterminant de la matrice jacobienne sérielle **B** est fonction, comme pour la matrice jacobienne parallèle **A**, de la configuration articulaire **q** et de la configuration de la plate-forme **X**. Pour une position de l'organe effecteur, il est possible d'associer 4 solutions du modèle géométrique inverse. On calcule la valeur du déterminant de la matrice jacobienne sérielle **B** pour toutes ces solutions, c'est-à-dire pour tous les modes de fonctionnement de ce manipulateur.

Les modes de fonctionnement (Définition 7) permettent de séparer facilement ces solutions car ils sont définis à partir de la matrice jacobienne sérielle.

Le manipulateur RR-RRR possède deux jambes indépendantes, on peut donc associer deux modes de fonctionnement pour chaque jambe (Figure 8) pour une même position du point $C_1$. Ainsi, il existe 4 modes de fonctionnement ($Mf_1$, $Mf_2$, $Mf_3$, $Mf_4$) définis dans le tableau 1.

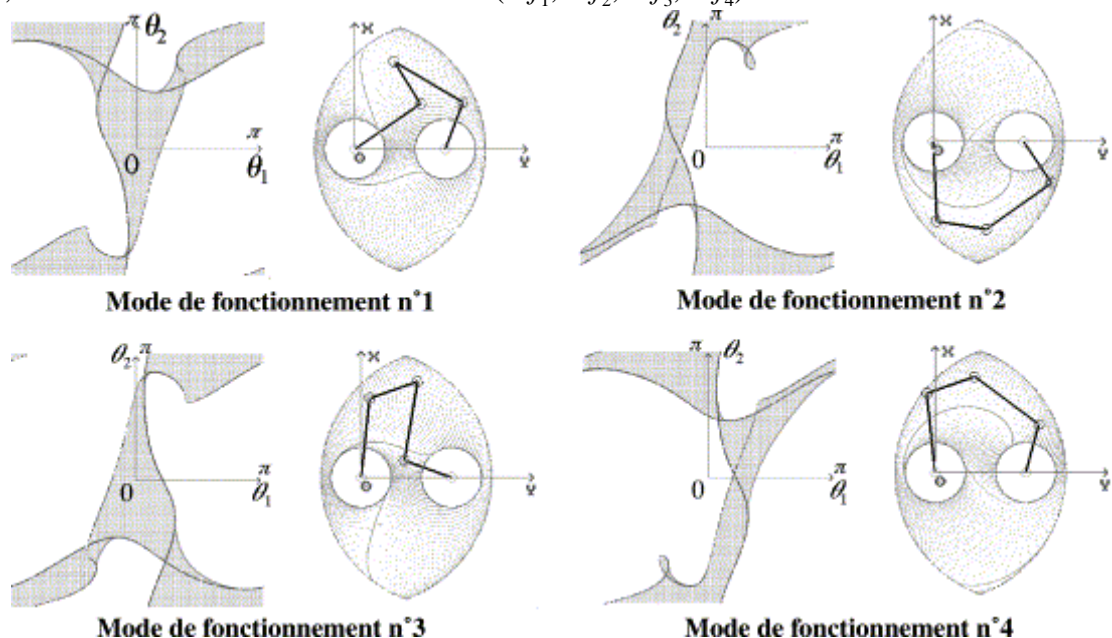

**Figure 8 : Les 4 modes de fonctionnement du manipulateur RR-RRR**

D. Chablat et Ph. Wenger     Journal of Mechanism and Machine Theory, Vol 36/6, pp. 763-783, 2001

| $B_{ii}$ | $Mf_1$ | $Mf_2$ | $Mf_3$ | $Mf_4$ |
|---|---|---|---|---|
| $B_{11}=L_1\sin(\theta_1)\ (x - L_1\cos(\theta_1)) - L_1\cos(\theta_1)\ (y - L_1\sin(\theta_1))$ | positif | positif | négatif | négatif |
| $B_{22}=L_2\sin(\theta_2)\ (x - L_2\cos(\theta_2) - c_2) - L_2\cos(\theta_2)\ (y - L_2\sin(\theta_2))$ | positif | négatif | négatif | positif |

**Tableau 1 : Les 4 modes de fonctionnement d'un manipulateur <u>R</u>R-<u>R</u>RR**

On en déduit que les aspects définis par Borrel (Définition 6) ne peuvent pas diviser l'ensemble articulaire en domaines disjoints car, pour le manipulateur parallèle <u>R</u>R-<u>R</u>RR, les domaines ainsi définis se recouvrent partiellement. Ils ne sont donc pas les plus grands domaines sans singularité.

### 3.3.3. Conclusion

Nous avons vu que les deux définitions des aspects ne peuvent s'appliquer aux manipulateurs pleinement parallèles avec plusieurs solutions aux modèles géométriques direct et inverse, car aucune n'est capable de séparer ces solutions, ni dans l'ensemble articulaire, ni dans l'espace de travail. Nous allons donc définir les aspects de ces manipulateurs dans le produit cartésien de l'espace de travail et de l'ensemble articulaire pour trouver les domaines exempts de toute configuration singulière.

### 3.4. LES ASPECTS : DEFINITION GENERALE

Dans les paragraphes 3.3.1 et 3.3.2, nous avons vu qu'il n'était pas possible de définir les aspects uniquement dans l'espace de travail ou dans l'ensemble articulaire pour les manipulateurs pleinement parallèles possédant plusieurs solutions à leurs modèles géométriques direct et inverse. À l'aide des modes de fonctionnement, nous allons définir les aspects pour ces manipulateurs que l'on nommera aspects généralisés.

**Définition 8 : Les aspects généralisés**

On appelle aspects généralisés $A_{ij}$ d'un manipulateur pleinement parallèle, les plus grands domaines connexes de $W \times Q$ tels que :

- $A_{ij} \subset W \times Q$ ;
- $A_{ij}$ est connexe ;
- $A_{ij} = \{(\mathbf{X}, \mathbf{q}) \in Mf_i \text{ tel que } \det(\mathbf{A}) \neq 0\}$

La projection $\pi_W$ des aspects généralisés sur l'espace de travail donne les domaines $WA_{i\,j}$ tels que :

- $WA_{i\,j} \subset W$ ;
- $WA_{i\,j}$ est connexe.

Ces domaines sont appelés, les W-aspects, ils sont les plus grands domaines connexes de l'espace de travail, pour un mode de fonctionnement donné $Mf_i$, exempts de singularité sérielle et parallèle.

La projection $\pi_Q$ des aspects généralisés sur l'ensemble articulaire donne les domaines tels que :

- $QA_{i\,j} \subset Q$ ;
- $QA_{i\,j}$ est connexe.

Ces domaines sont appelés, les Q-aspects, ils sont les plus grands domaines connexes de l'espace articulaire, pour un mode de fonctionnement donné, exempts de singularité sérielle et parallèle.

**Les aspects généralisés : Application au manipulateur RR-RRR**

Nous représentons donc les projections sur l'espace de travail et l'ensemble articulaire pour chaque mode de fonctionnement $Mf_i$. L'ensemble des résultats est synthétisé dans le tableau 2 et pour chaque mode de fonctionnement, nous trouvons le nombre d'aspects généralisés associés (Figure 9).

|  | **1** | **2** | **3** | **4+5** | **6** | **7+8** | **9** | **10** |
|---|---|---|---|---|---|---|---|---|
| det(**A**) | positif | positif | positif | positif | négatif | négatif | négatif | négatif |
| $B_{11}$ | positif | positif | négatif | négatif | positif | positif | négatif | négatif |
| $B_{22}$ | positif | négatif | négatif | positif | positif | négatif | négatif | positif |
| Nombre d'aspects généralisés | 1 | 1 | 1 | 2 | 1 | 2 | 1 | 1 |

**Tableau 2 : Définitions des aspects généralisés du manipulateur <u>RR</u> - <u>RRR</u>**

Les aspects généralisés sont les plus grands domaines exempts de toutes singularités. Pour notre exemple, ils permettent de séparer les solutions des modèles géométriques inverse et direct. Cependant, nous allons voir que cette propriété n'est pas vérifiée pour tous les manipulateurs pleinement parallèles.

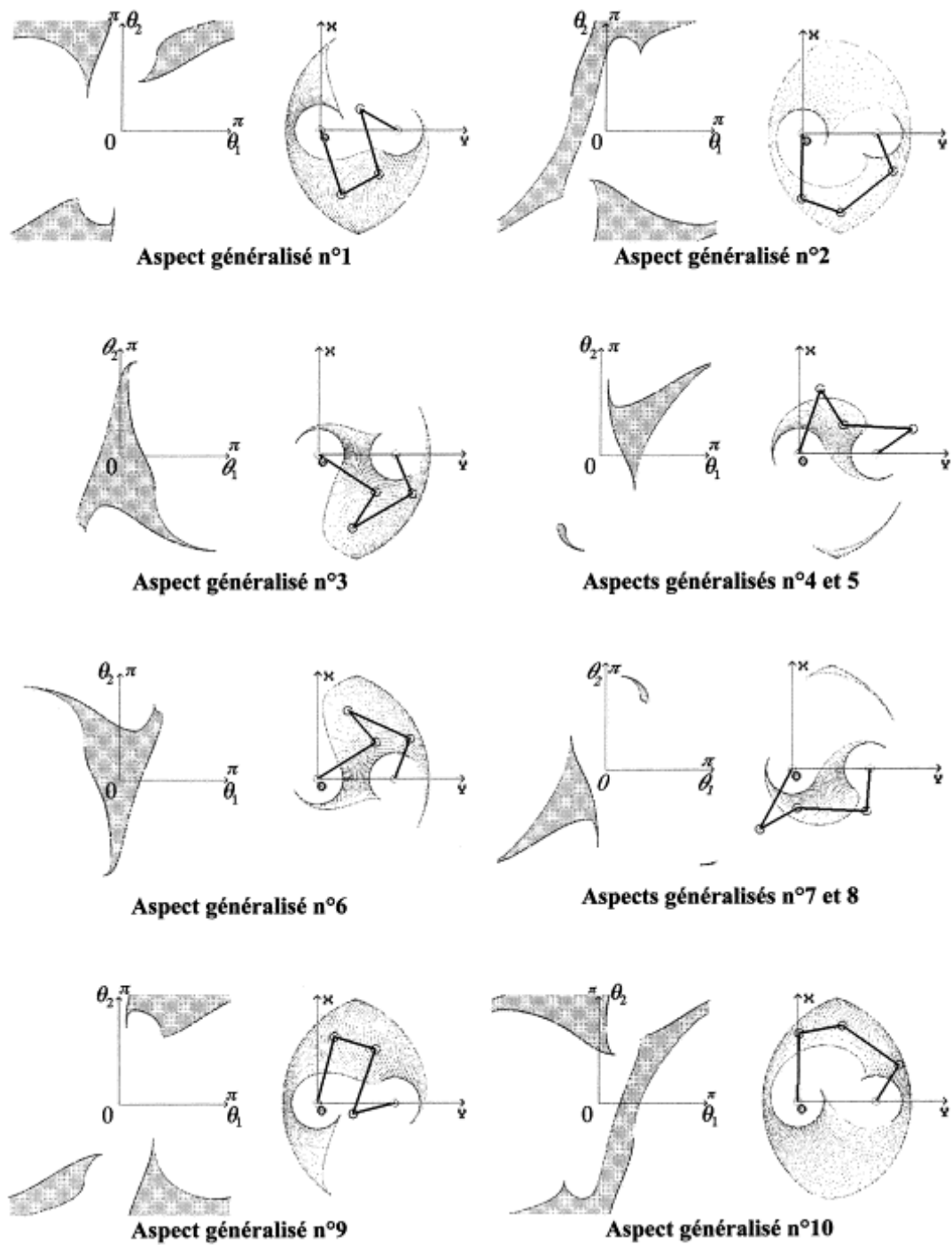

**Figure 9 : Les aspects généralisés du manipulateur <u>R</u>R-<u>R</u>RR**

## 4. LES DOMAINES D'UNICITE

### 4.1. ILLUSTRATION DU PROBLEME A L'AIDE D'UN MANIPULATEUR A TROIS DEGRES DE LIBERTE

Nous avons choisi le manipulateur 3-RRR qui est étudié par **[13]**, **[6]**. Il possède plusieurs solutions aux modèles géométriques inverse et direct. Sa morphologie est proche de celle du manipulateur de type 3-RPR étudié par **[3]** pour lequel il est montré qu'il est possible de changer de mode d'assemblage sans passer par une singularité.

#### 4.1.1. Le modèle géométrique inverse

L'équation **12** permet la résolution du modèle géométrique inverse de ce manipulateur.

$$\begin{pmatrix} (x + C_{i\,x} \cos(\phi) - C_{i\,y} \sin(\phi) - A_{i\,x} - L_i\cos(\alpha_i))^2 + \\ (y + C_{i\,x} \sin(\phi) + C_{i\,y} \cos(\phi) - A_{i\,y} - L_i\sin(\alpha_i))^2 \end{pmatrix} = M_i^2 \quad \text{pour i = 1, 2, 3} \quad (12)$$

où $A_{i\,x}$ et $A_{i\,y}$ sont les cordonnées des points $A_i$ dans le repère fixe et $C_{i\,x}$ et $C_{i\,y}$ sont les cordonnées des points $C_i$ dans le repère lié à la plate-forme mobile.

#### 4.1.2. Le modèle géométrique direct

La résolution de ce problème est abordée par **[15]**. Nous pouvons résoudre le problème du modèle géométrique direct de ce manipulateur, en utilisant les résultats obtenus pour le manipulateur à boucle fermée de type 3 - RPR **7]**. Nous calculons les coordonnées des points $C_i$ à partir de la méthode de résolution du manipulateur 3 - RPR, en fixant les longueurs des articulations prismatiques égales aux longueurs des segments $M_i$. Nous avons donc le même nombre de solutions que pour le manipulateur 3 - RPR, c'est-à-dire 6.

#### 4.1.3. Singularités

**Étude des singularités parallèles**

Les singularités parallèles se produisent lorsque det(**A**) = 0.

$$\frac{\partial \begin{pmatrix} (x + C_{i\,x} \cos(\phi) - C_{i\,y} \sin(\phi) - A_{i\,x} - L_i\cos(\alpha_i))^2 + \\ (y + C_{i\,x} \sin(\phi) + C_{i\,y} \cos(\phi) - A_{i\,y} - L_i\sin(\alpha_i))^2 \end{pmatrix}}{\partial x} = P_{i\,1} \quad \text{pour i = 1, 2, 3.} \quad (13a)$$

$$\frac{\partial \begin{pmatrix} (x + C_{i\,x} \cos(\phi) - C_{i\,y} \sin(\phi) - A_{i\,x} - L_i\cos(\alpha_i))^2 + \\ (y + C_{i\,x} \sin(\phi) + C_{i\,y} \cos(\phi) - A_{i\,y} - L_i\sin(\alpha_i))^2 \end{pmatrix}}{\partial y} = P_{i\,2} \quad \text{pour i = 1, 2, 3.} \quad (13b)$$

$$\frac{\partial \begin{pmatrix} (x + C_{ix}\cos(\phi) - C_{iy}\sin(\phi) - A_{ix} - L_i\cos(\alpha_i))^2 + \\ (y + C_{ix}\sin(\phi) + C_{iy}\cos(\phi) - A_{iy} - L_i\sin(\alpha_i))^2 \end{pmatrix}}{\partial \phi} = P_{i3} \text{ pour i = 1, 2, 3.} \quad (13c)$$

En utilisant les équations 13, nous pouvons ainsi écrire det(**A**) :

$$\det(\mathbf{A}) = \begin{vmatrix} P_{11} & P_{12} & P_{13} \\ P_{21} & P_{22} & P_{23} \\ P_{31} & P_{32} & P_{33} \end{vmatrix} \quad (14)$$

Le déterminant de la matrice jacobienne parallèle **A** est fonction des variables d'entrée $\theta$ c'est-à-dire de $[\alpha_1, \alpha_2, \alpha_3]^T$. Les singularités parallèles sont les configurations de la plate-forme mobile pour lesquelles les droites $B_iC_i$ ont un point commun.

**Étude des singularités sérielles**

Les singularités sérielles se produisent lorsque det(**B**) = 0.

$$\frac{\partial \begin{pmatrix} (x + C_{ix}\cos(\phi) - C_{iy}\sin(\phi) - A_{ix} - L_i\cos(\alpha_i))^2 + \\ (y + C_{ix}\sin(\phi) + C_{iy}\cos(\phi) - A_{iy} - L_i\sin(\alpha_i))^2 \end{pmatrix}}{\partial \alpha_i} =$$

$$2L_i \begin{pmatrix} \sin(\alpha_i)(x + C_{ix}\cos(\phi) - C_{iy}\sin(\phi) - A_{ix}) + \\ \cos(\alpha_i)(y + C_{ix}\sin(\phi) + C_{iy}\cos(\phi) - A_{iy}) \end{pmatrix} = B_{ii} \text{ pour i = 1, 2, 3}$$

Nous pouvons ainsi écrire l'expression de det(**B**) :

$$\det(\mathbf{B}) = \prod_1^3 2L_i \begin{pmatrix} \sin(\alpha_i)(x + C_ix\cos(\phi) - C_iy\sin(\phi) - A_ix) - \\ \cos(\alpha_i)(y + C_ix\sin(\phi) + C_iy\cos(\phi) - A_iy) \end{pmatrix}$$

Pour interpréter les configurations singulières de ce type, nous posons les expressions suivantes :

$$\overrightarrow{A_iC_i} = (x + C_ix\cos(\phi) - C_iy\sin(\phi) - A_ix,\ y + C_ix\sin(\phi) + C_iy\cos(\phi) - A_iy) \quad \text{et}$$

$$\overrightarrow{A_iB_i} = L_i(\sin(\alpha_i), \cos(\alpha_i))$$

Chaque facteur du déterminant s'annule lorsque $\overrightarrow{A_iC_i} \wedge \overrightarrow{A_iB_i} = \vec{0}$, c'est-à-dire lorsque les points $A_i$, $B_i$ et $C_i$ sont alignés, pour i = 1, 2, 3.

### 4.1.4. Les aspects généralisés

Nous définissons les aspects généralisés de ce manipulateur à partir de la définition 8. Pour les dimensions du manipulateur étudié (Tableau 3), nous avons ainsi 8 modes de fonctionnement.

| $A_{1x}$ = -10,0 | $A_{2x}$ = 10,0 | $A_{3x}$ = 0,0 |
|---|---|---|
| $A_{1y}$ = -10,0 | $A_{2y}$ = -10,0 | $A_{3y}$ = 10,0 |
| $C_{1x}$ = 0,0 | $C_{2x}$ = 10,0 | $C_{3x}$ = 10,0 |
| $C_{1y}$ = 0,0 | $C_{2y}$ = 0,0 | $C_{3y}$ = 10,0 |
| $L_1$ = 10,0 | $L_2$ = 10,0 | $L_3$ = 10,0 |
| $M_1$ = 10,0 | $M_2$ = 10,0 | $M_3$ = 10,0 |

**Tableau 3 : Dimensions d'un manipulateur parallèle 3-RRR**

Il n'est pas possible de représenter ici tous les aspects généralisés de ce manipulateur. Cependant, nous avons trouvé des aspects parallèles possédant plusieurs solutions au modèle géométrique direct, ce qui nous permet de dire que la notion d'aspect généralisé n'est pas suffisante pour déterminer les domaines d'unicité de ces manipulateurs.

### 4.1.5. Trajectoire de changement de mode d'assemblage

Nous montrons ici qu'il est possible de réaliser une trajectoire permettant de changer de mode d'assemblage sans passer ni par une singularité parallèle ni par une singularité sérielle. Nous définissons cette trajectoire à partir de la modélisation d'un aspect généralisé $A_{ij}$.

La trajectoire réalisée est une interpolation linéaire dans l'espace de travail entre les configurations de la plate-forme mobile données dans le tableau suivant :

|  | X | Y | $\phi$ (rad) |
|---|---|---|---|
| Point 1 | -15,468 | 0,781 | 0,073 |
| Point 2 | -15,468 | -7,091 | 0,500 |
| Point 3 | -9,902 | -7,091 | 1,081 |

**Tableau 4 : Trajectoire de changement de mode d'assemblage**

La configuration de départ et la configuration d'arrivée de cette trajectoire, ainsi qu'une configuration intermédiaire sont représentées dans la figure 10.

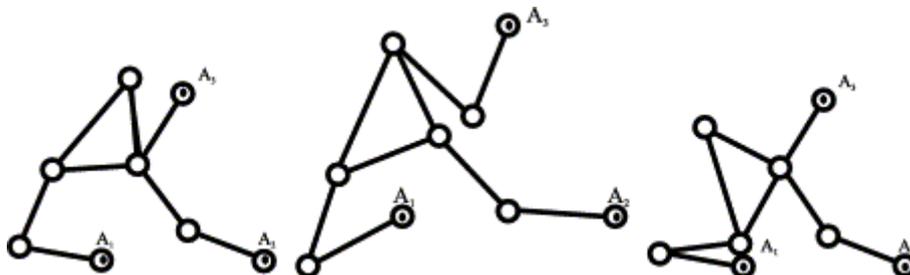

**Figure 10 : Changement de mode d'assemblage pour un manipulateur parallèle 3-RRR**

Nous avons calculé la valeur des différents déterminants (**A** et **B**) durant cette trajectoire. Nous pouvons ainsi vérifier qu'aucun d'entre eux ne s'annule (Figure 11 : En ordonnée, nous avons normé les valeurs de Det(**A**), $B_{11}$, $B_{22}$, $B_{33}$).

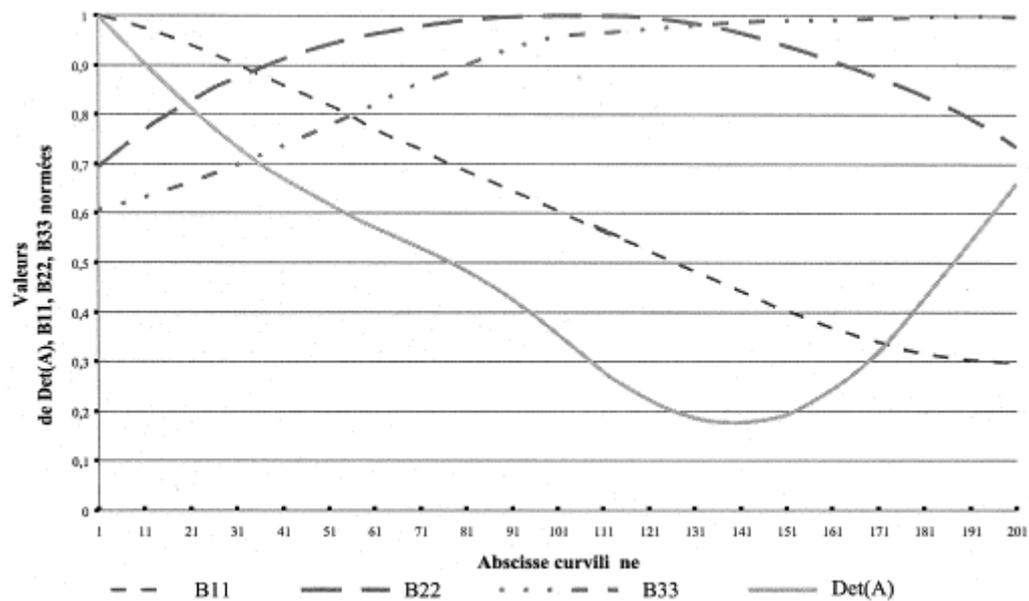

**Figure 11 : Variation de la valeur des déterminants le long de la trajectoire**

### 4.2. LES SURFACES CARACTERISTIQUES

Les surfaces caractéristiques ont été introduites par **[2]** pour redéfinir les domaines d'unicité des robots sériels de morphologie cuspidale. Cette définition a été étendue pour la famille des manipulateurs pleinement parallèles possédant une seule solution au modèle géométrique inverse **[3]**.

Nous allons généraliser cette définition aux manipulateurs pleinement parallèles possédant plusieurs solutions au modèle géométrique inverse.

**Définition 9 : Les surfaces caractéristiques**

Soit $WA_{ij}$ un aspect parallèle de l'espace de travail $W$. On définit les *surfaces caractéristiques* de l'aspect parallèle $WA_{ij}$, noté $S_C(WA_{ij})$, comme l'image réciproque dans $WA_{ij}$ de l'image des frontières $\overline{WA_{ij}}$ qui délimitent $WA_{ij}$.

$$S_C(WA_{ij}) = g^{-1,i}\left(g_i(\overline{WA_{ij}})\right) \cap WA_{ij} \qquad (15)$$

où $g_i$ et $g^{-1,i}$ sont définies respectivement avec les équations (5) et (6).

La figure 12 montre le principe d'obtention des surfaces caractéristiques. Pour chaque aspect généralisé, on considère leurs frontières qui correspondent aux configurations singulières et aux butées articulaires. En calculant sur cette frontière le modèle géométrique inverse, on obtient une configuration articulaire **q**. Puis, à partir de cette configuration **q**, on résout le modèle géométrique direct. On obtient naturellement la configuration de la plate-forme mobile initiale (configuration singulière) mais un ensemble d'autres configurations régulières. Pour former une surface caractéristique, nous réunissons l'ensemble de ces configurations appartenant à l'aspect initial.

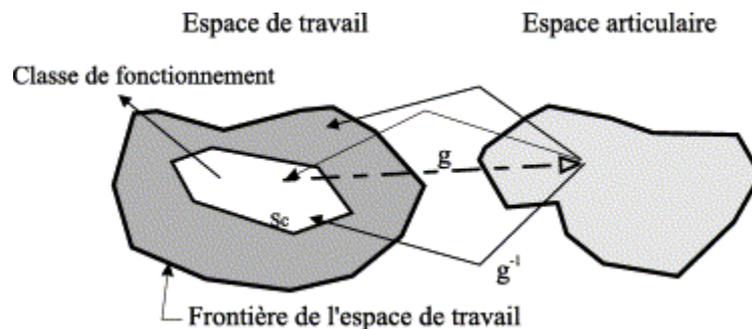
**Figure 12 : Construction des surfaces caractéristiques**

**Remarque :**

Cette définition suppose que les jambes du manipulateur ne sont pas cuspidales. Dans le cas contraire, il est nécessaire de définir des surfaces caractéristiques dans l'ensemble articulaire pour séparer les solutions du modèle géométrique inverse.

### 4.3. LES REGIONS DE BASE ET LES COMPOSANTES DE BASE

Le but des surfaces caractéristiques est de subdiviser les aspects pour séparer les solutions du modèle géométrique direct. Les domaines ainsi obtenus seront les *appelés régions* de base dans l'espace de travail et leur image dans l'ensemble articulaire, les *composantes de base*.

**Définition 10 : Régions de base**

Soit $WA_{ij}$ un W-aspect d'un mode de fonctionnement $Mf_i$ appartenant à l'espace de travail $W$. On définit les *régions de base*, notées $WAb_{ijk}$, ($k \in K$ ensemble des régions de base) comme étant les composantes connexes de $WA_{ij} \setminus S_C(WA_{ij})$ ($\setminus$ désigne la différence entre ensembles). Les régions de base réalisent une partition de $WA_{ij}$ :

$$WA_{ij} = (\cup_{k \in K} WA_{ijk}) \, S_C(WA_{ij}) \tag{16}$$

L'espace de travail est donc découpé par les singularités en aspects généralisés, et eux-mêmes subdivisés en régions de base à l'aide des surfaces caractéristiques (Figure 13).

D. Chablat et Ph. Wenger	Journal of Mechanism and Machine Theory, Vol 36/6, pp. 763-783, 2001

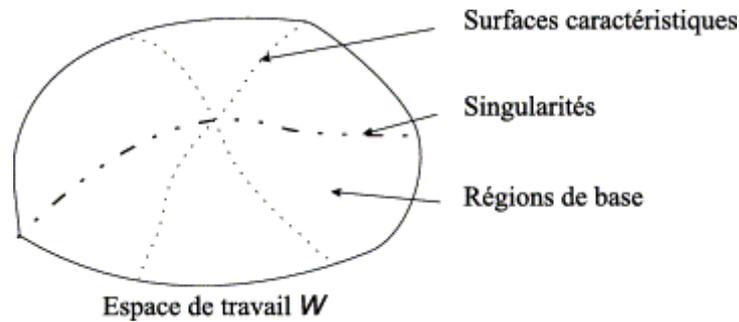

**Figure 13 : Décomposition de l'espace de travail en régions de base**

**Définition 11 : Composantes de base**

Soit $QAb_{ijk} = g_i(WAb_{ijk})$, $QAb_{ijk}$ est un domaine de l'ensemble articulaire Q qu'on appelle *composantes de base*.

Soit $WA_{ij}$ un W-aspect et $QA_{ij}$ son image par $g_i$, on a :

$$QA_{ij} = (\cup_{k \in K} QAb_{ijk}) \cup g_i(S_C(WA_{ij})) \qquad (17)$$

**Théorème 2 :**

En d'autres termes, les régions de base sont les domaines d'unicité pour le modèle géométrique direct.

**Conséquence 1 :**

Les composantes de base images des régions de base d'un même aspect généralisé, pour un mode de fonctionnement donné, sont soient disjointes, soient confondues.

La démonstration de ce théorème est analogique à celle donnée dans **[7]** pour le cas des manipulateurs parallèles possédant une seule solution au modèle géométrique inverse.

Le nombre de composantes de base confondues est fonction du nombre de solutions au modèle géométrique direct. La figure 14 symbolise la décomposition de l'ensemble articulaire en composantes de base.

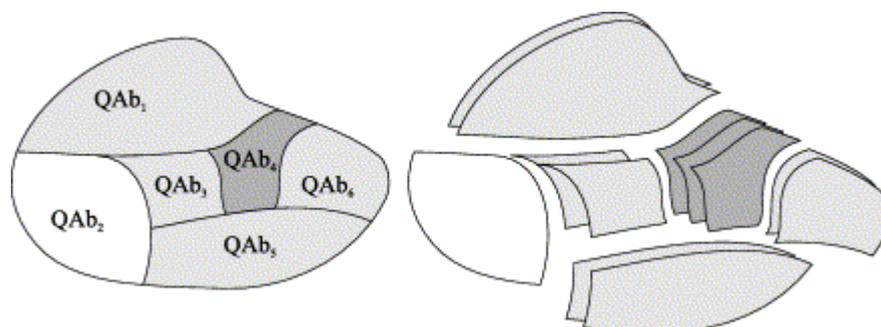

**Figure 14 : Décomposition de l'ensemble articulaire en composantes de base**

### 4.4. LES DOMAINES D'UNICITE : DEFINITION GENERALE

Le théorème 2 permet de caractériser les domaines de l'espace de travail possédant une seule solution au modèle géométrique direct. Le théorème 3 suivant a pour but de définir les domaines d'unicité les plus grands de l'opérateur géométrique : tout autre domaine d'unicité est alors inclus dans l'un d'eux.

**Théorème 3 :**

Les domaines d'unicité $Wu_{i\,k}$ sont les domaines obtenus comme la réunion de l'ensemble $(\cup_{j \in J'} WAb_{i\,j})$ des régions de base adjacentes d'un même W-aspect $WA_{i\,j}$ dont les images respectives par $g_i$ sont disjointes, et du sous-ensemble $S_C(J')$ des surfaces caractéristiques qui séparent ces composantes de base :

$$Wu_{i\,k} = (\cup_{j \in J'} WAb_{i\,j}) \cup S_C(J') \qquad (18)$$

avec $J' \subset J$ tel que $\forall j_1, j_2 \in J'$, $g_i(WAb_{i\,j_1}) \cap g_i(WAb_{i\,j_2}) = \varnothing$

La figure 15 permet de schématiser la démarche utilisée pour construire les domaines d'unicité les plus grands. Nous créons deux domaines d'unicité $Wu_{1\,1}$ et $Wu_{1\,2}$ avec $Wu_{1\,1} = WAb_{1\,1} \cup WAb_{1\,2}$ et $Wu_{1\,2} = WAb_{1\,2} \cup WAb_{1\,3}$. Il n'est pas possible de construire un domaine d'unicité comportant les composantes $WAb_{1\,1}$, $WAb_{1\,2}$ et $WAb_{1\,3}$ car les images de $WAb_{1\,1}$ et de $WAb_{1\,3}$ dans l'ensemble articulaire sont confondues. De la même manière, il n'est pas possible d'associer $WAb_{1\,4}$ à $Wu_{1\,1}$ ou $Wu_{1\,2}$ car son image n'est adjacente à aucune autre composante de base. Dans ce cas, $WAb_{1\,4}$ constitue un troisième domaine d'unicité $Wu_{1\,3}$.

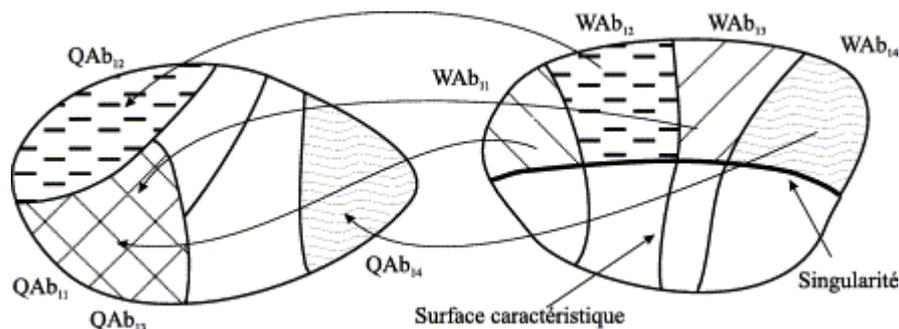

**Figure 15 : Les domaines d'unicité**

## 5. CONCLUSION

Nous avons introduit les modes de fonctionnement pour séparer les solutions du modèle géométrique inverse. Puis, nous avons généralisé la notion d'aspect à l'ensemble des manipulateurs pleinement parallèles. À l'intérieur de ces aspects, nous n'avons pas de singularité parallèle ni de singularité sérielle. Cependant, nous avons montré que cette notion est insuffisante pour déterminer les domaines d'unicité de tous ces manipulateurs. Nous avons ainsi montré qu'il est possible de changer de mode d'assemblage sans passer par une singularité parallèle ni une singularité sérielle. Aussi, nous avons étendu la définition des surfaces caractéristiques, des régions de base et des composantes de base pour ces manipulateurs. Ces définitions sont une extension des définitions utilisées pour les manipulateurs pleinement parallèles possédant une seule solution à leur modèle géométrique inverse **[3]**. Ces nouveaux domaines d'unicité vont ainsi nous permettre de formaliser les notions de parcourabilité.